\documentclass[runningheads]{llncs}

\usepackage{eccv}

\usepackage{eccvabbrv}

\usepackage{graphicx}
\usepackage{booktabs}

\usepackage[accsupp]{axessibility}  

\usepackage{hyperref}

\usepackage{orcidlink}

\begin{document}

\title{Connectivity-Inspired Network for Context-Aware Recognition} 

\titlerunning{Connectivity-Inspired Network for Context-Aware Recognition}

\author{
Gianluca Carloni\inst{1,2}\orcidlink{0000-0002-5774-361X} \and
Sara Colantonio\inst{1}\orcidlink{0000-0003-2022-0804}
}

\authorrunning{G.~Carloni \etal}

\institute{
National Research Council of Italy, Via Moruzzi 1, 56127 Pisa, Italy
\email{gianluca.carloni@isti.cnr.it}\\
\and
University of Pisa, Via Caruso 16, 56122 Pisa, Italy 
}

\maketitle

\begin{abstract}
  The aim of this paper is threefold. We inform the AI practitioner about the human visual system with an extensive literature review; we propose a novel biologically motivated neural network for image classification; and, finally, we present a new plug-and-play module to model context awareness. We focus on the effect of incorporating circuit motifs found in biological brains to address visual recognition. Our convolutional architecture is inspired by the connectivity of human cortical and subcortical streams, and we implement bottom-up and top-down modulations that mimic the extensive afferent and efferent connections between visual and cognitive areas. Our Contextual Attention Block is simple and effective and can be integrated with any feed-forward neural network. It infers weights that multiply the feature maps according to their causal influence on the scene, modeling the co-occurrence of different objects in the image. We place our module at different bottlenecks to infuse a hierarchical context awareness into the model. We validated our proposals through image classification experiments on benchmark data and found a consistent improvement in performance and the robustness of the produced explanations via class activation. Our code is available at  \url{https://github.com/gianlucarloni/CoCoReco}. 
\end{abstract}

\section{Introduction}
\label{sec:intro_cocoreco}
The aim of this paper is threefold and we make the following main contributions. \textbf{1)} Informing the AI practitioner about the human visual system. We review fundamental notions and recent trends in the study of human vision - what are the ventral and dorsal streams and how they communicate, how top-down modulation occurs, the existence of subcortical pathways, and the importance of context in vision. This can foster human-inspired computer vision.
\textbf{2)} Proposing a novel biologically motivated neural network for image classification. We design a convolutional model conceptually inspired by the above-mentioned mechanism of human vision and numerically based on recent connectomic studies.
\textbf{3)} Presenting a new plug-and-play module to model context awareness. Our contextual attention block (CAB) can be added to any traditional feed-forward architecture to improve recognition by modelling feature co-occurence in the real world.

\section{Related Work}
\subsection{The Human Visual System}
Historically, visual information processing in humans was described with the \textit{two-streams theory}, which involved the existence of two anatomically distinct and functionally specialized cortical pathways, the \textbf{ventral stream}, which processes visual features like color, size, and dimension, and the \textbf{dorsal} stream, which primarily deals with the object's spatial features (location, orientation, and motion) \cite{goodale1992separate,whitwell2014two}.
\begin{figure}[tb]
  \centering
  \includegraphics[width=0.65\textwidth]{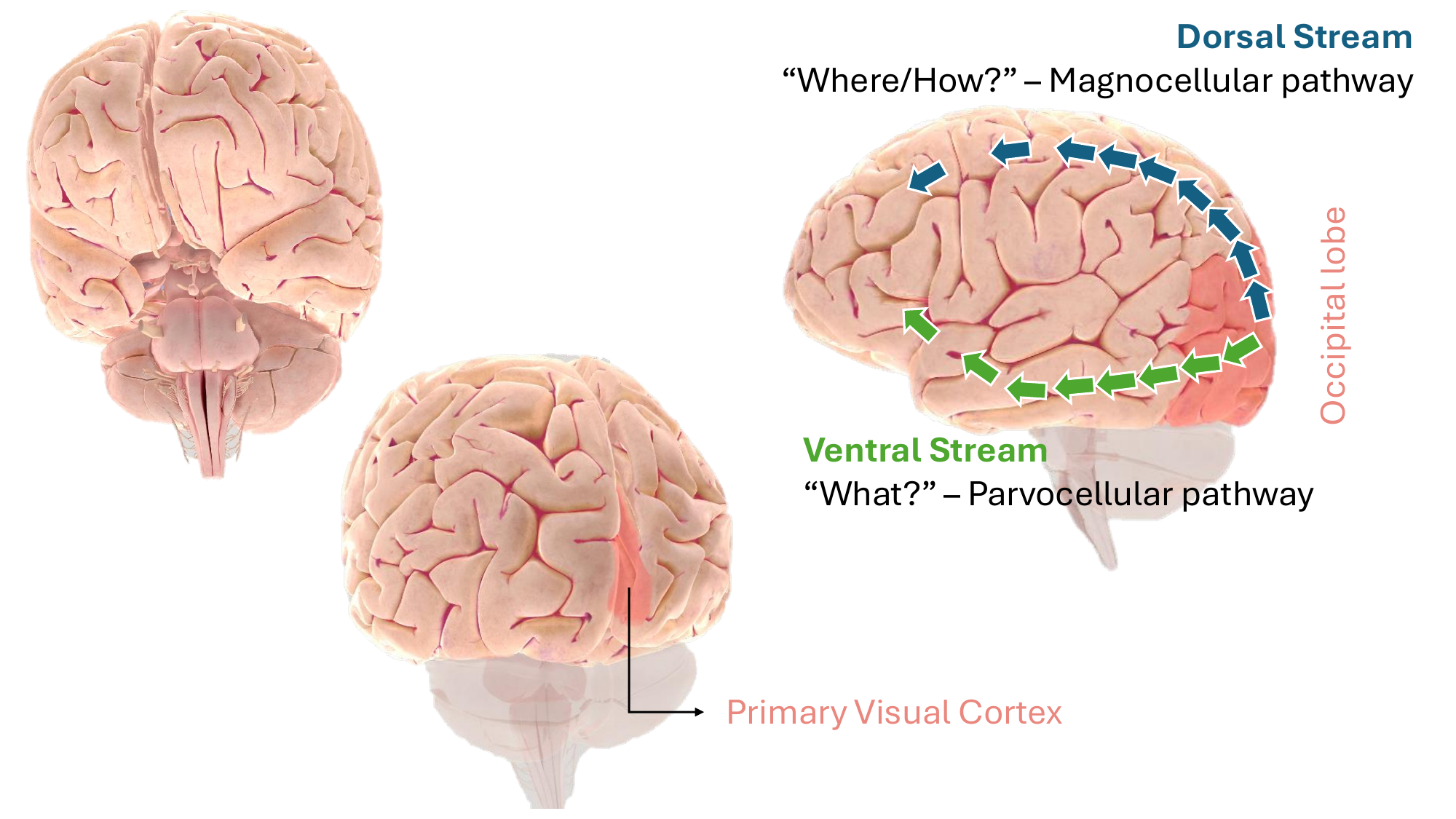}
  \caption{The ventral and dorsal visual pathways in human vision. The brain models depicted in this image are adapted from https://www.brainfacts.org/ of the Society for Neuroscience (2017).}
  \label{fig:cab_module}
\end{figure}
The \textbf{ventral} stream runs from the primary visual cortex (V1) to the temporal lobe, mainly through extrastriate visual areas II (V2) and IV (V4), to the inferotemporal cortex (IT or ITC), and then to the prefrontal cortex (PFC), which is involved in linking perception to memory and action. This pathway is responsible for object recognition, categorization, and memorization, thus representing object shape and identity (i.e., the \textit{What?} of a visual scene).
Since it receives signals from the parvocellular cell (P cell) layers, which are sensitive to color and have a higher spatial resolution but lower temporal resolution, the ventral pathway processes the feature-rich information for fine local processing (i.e., textures and edges) in a bottom-up manner to form detailed representations of visual stimuli.
Indeed, cortical visual areas from V1 to ITC form an anatomical hierarchy in which information is processed sequentially with increasing complexities \cite{konen2008two}, and the invariance of those representations to position and scale increase. In parallel, there is also an increase in the size of the receptive fields as well as in the complexity of the optimal stimuli for the neurons \cite{gattass1981visual,gattass1988visuotopic}.

The \textbf{dorsal} stream is responsible for spatial perception, motion detection and attention, and thus represents spatial vision and visuomotor control. Indeed, it represents object location or spatial relationships (\textit{Where?}) within the visual stimuli.
Anatomically and functionally, the dorsal stream runs from V1 to the parietal lobe. Since it receives retinal information from the magnocellular (M cell) layers, it involves fast processing of the information sensitive to luminance changes (high contrast gain) and low spatial frequencies (LSF).
Whereas visual representations in the ventral pathway are more invariant and reflect \textit{“what an object is”}, those in the dorsal pathway are more adaptive and reflect \textit{“what we do with it”}. In this sense, the dorsal stream is addressed as the vision-for-\textit{action} path, in contrast to the vision-for-\textit{perception} representations in the ventral stream. 
At some level of neural processing, information about the identity ('what': ventral) and location ('where'/'how': dorsal) of an object represented in the segregated pathways must be integrated. To this end, many schemes have been proposed in the literature. As reviewing the solutions proposed in recent years is outside the scope of this work, we point the reader to some examples like \cite{perry2014feature}, \cite{choi2020proposal}, and \cite{milleret2018beyond} to show how no strict consensus on the ventral-dorsal modeling has been reached yet.

\subsection{Ventral and Dorsal Streams Communicate}
If the general belief in the 1970s to the 1990s was that of a complete division of labor by two segregated visual pathways, this started to change in the early 2000s. New evidence began to emerge calling for synergies between them: there exist “what” and “where” information in \textit{both} visual processing pathways.
Evidence supports the hypothesis that \textbf{dorsal and ventral visual areas communicate}, and there are shape-selectivity and non-action-based perceptual representations in the posterior regions of the dorsal pathway \cite{lehky2007comparison,konen2008two,takemura2016major}. Moreover, the dorsal pathway itself is composed of several sub-pathways, where at least one has a functional, and probably necessary, role in object perception \cite{rousselet2004parallel}. 
Under this light, the two streams are not segregated but constitute an important symbiosis crucial in transmitting signals between regions. Shape encoding is thus performed \textit{also} in the dorsal pathway, and it is distinct from and not a mere duplication of that formed in the ventral pathway. This means that dorsal object representations are dissociable from those generated in the ventral pathway and play an independent and functional role in visual perception.
Therefore, visual perception should be studied not simply as a function of one (ventral) "what" pathway, but rather as the joint outcome of the processing and coordination of different "what" regions in both cortical visual pathways.

\subsection{A Fast Top-Down Modulation of Bottom-Up Representations}
In addition to the traditional bottom-up hierarchy of representation, new mechanisms of top-down processing were proposed \cite{bar2001cortical,bar2003cortical,bar2006top}. There exists a non-cortical, fast, "shortcut" stream in which early visual inputs are sent, partially analyzed, from the early visual cortex  (V1) to the prefrontal cortex (PFC). Possible interpretations of the crude visual input are generated in the PFC and then sent to the inferotemporal cortex (IT/ITC), subsequently activating relevant object representations, which are then incorporated into the slower, bottom-up process \cite{bar2006very,bar2007proactive}. Eventually, the coarse and global representations from the subcortical pathway guide and modulate the fine local representations from the ventral pathway in a \textbf{“top-down”} fashion to form more precise representations of the object category.
In this view, prefrontal cortex (PFC) areas would receive coarse, low-resolution information via fast dorsal projections and generate predictions about object identity. This prediction would be feedback to the temporal cortex, facilitating recognition by limiting the number of possible object candidates \cite{paneri2017top}.
The key insight is that this “faster” subcortical pathway is parallel (or completed prior) to the ventral pathway.

\subsection{Lateral Pathways - Superior Colliculus and Pulvinar}
In humans as well as other mammals, the two strongest pathways linking the eye to the brain are those projecting to the dorsal part of the LGN in the thalamus and to the superior colliculus (SC)\cite{goodale2013sight}.
From the former originate the well-known schematic of ventral and dorsal streams. At the same time, the latter constitutes the other major retino–cortical visual pathway known as the tectopulvinar pathway, routing primarily through the superior colliculus (SC) and thalamic pulvinar (Pulv) nucleus onto ventral visual area V4 and dorsal visual area V5/MT.

\subsection{Visual Context in Object Recognition}
Context is of fundamental importance to both human and machine vision; e.g., an object in the air is more likely to be an airplane than a pig. The rich notion of context incorporates several aspects including physics rules, statistical co-occurrences, and relative object sizes, among others \cite{bomatter2021pigs}.
Indeed, context is of critical importance for locating a target object in complex scenes as it helps narrow down the search area and makes the search process more efficient \cite{ding2022efficient}. It is no surprise that AI and computer vision solutions embedding context information has emerged in the literature. Examples include zero-shot visual search \cite{ding2022efficient, bomatter2021pigs}, context-aware attention networks \cite{zhang2020putting}, 
and other computer vision models \cite{torralba2003contextbased}, \cite{choi2012context}, \cite{oliva2007role}, \cite{mack2011object}.
Unlike such solutions, as we shall see in the following, our proposal do not add computational overhead in terms of parameters, as it does not involve additional trainable parameters.

\section{Methods}
\subsection{Architecture design}
After reviewing the relevant literature, our second main contribution in this paper is the design of the connectivity-inspired context aware recognition network (\textbf{CoCoReco}). It is a dual-branched architecture for image classification inspired by the human ventral and dorsal streams and the tectopulvinar pathway. Moreover, we conceive a top-down modulation of the bottom-up representations from the pre-frontal cortex (PFC) and extensive afferent and efferent projections based on connectome studies.
As we shall see later, we placed our contextual attention block (CAB) at different bottlenecks to infuse a hierarchical context awareness into the model.
We design CoCoReco as a from-scratch ANN because evidence points to the opportunity to simplify ANNs to align with the visual streams better, with smaller and less complex ANNs being more brain-like than many of the best-performing ImageNet models \cite{schrimpf2018brain}.

Instead of modeling a single hierarchy of concentrical representations, we implemented a multi-branched convolutional architecture which considers that shape information is processed ubiquitously in different brain regions.
This way, we have operationalized a proper vision-for-perception schema based on both ventral and dorsal streams.
We designed each brain area as a 2D convolutional layer.
We route 90 percent of the retinal signal to the LGN layer and 10 percent to the SC and pulvinar layers, emulating the division of axons. Moreover, those layers are dominated by M cells. Thus, we model this faster and coarser information by increasing the kernel size and stride of the convolutions.

\begin{figure}[tb]
  \centering
  \includegraphics[width=\textwidth]{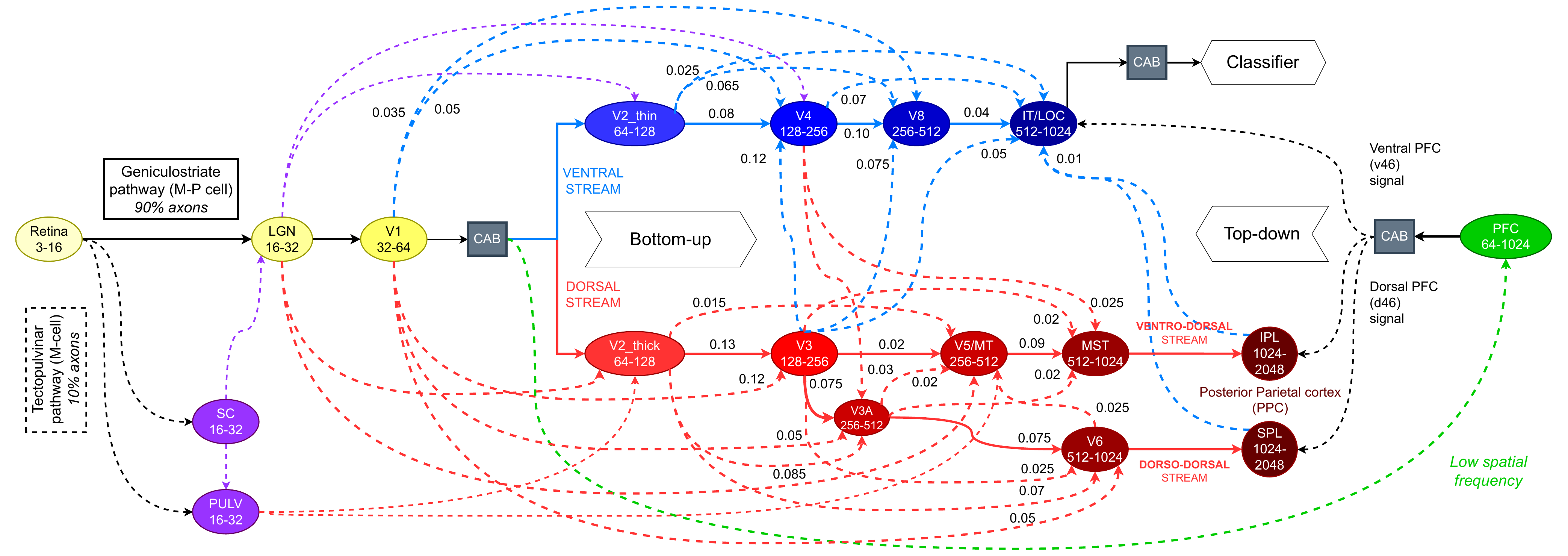}
  \caption{Overview of our \textbf{Co}nnectivity-inspired \textbf{Co}ntext aware \textbf{Reco}gnition network. The internals and rationale of CAB module is presented in Section 3.2 and Fig 3.
  Other abbreviations: lateral geniculostriate nucleus (LGN), superior colliculus (SC), pulvinar (PULV).
  Best seen in color.}
  \label{fig:cocoreco}
\end{figure}

To model the passing of information from one visual area to other areas, we design proper skip connections with a projection layer. The number and size of feature maps can be higher or lower depending on whether the information is conveyed in a forward (bottom-up) or backward (top-down) pass. For instance, evidence suggests that the V1 signal is transmitted not only to the visual area V2 (which directly follows V1 layer in the model), but also to later areas like V4 or V8, which have a more but smaller feature maps compared to V1, according to the hierarchy of constructed representations. To achieve information pass, we thus need to adjust the number and size of earlier representations. To this end, we design proper \textit{projection layers} made of a trainable convolutional layer adjusting the number of feature maps, followed by (i) a 2D average pooling, if it is a \textit{formward} pass, or (ii) a bilinear upsampling if it is a \textit{backward} pass. 

Recent studies on effective connectivity (EC) between brain regions reveal not only if two brain areas are anatomically/functionally connected but also expose the (relative) strength of such connections \cite{rolls2023multiple}. Therefore, we employ the EC measure found in that work as numerical estimates for the strength of the forward/feedback connections described above. As a result, we achieve a weighted projection by multiplying the output of the projection layers by the estimated weight.

\subsection{Contextual Attention Blocks}
As our third major contribution, we present a new plug-and-play module to inject context awareness into the model. In fact, our CoCoReco solution also models another fundamental aspect of human vision: context. To conceive our contextual attention block, CAB, we get inspiration from \cite{carloni2023causality,carloni2024exploiting} as a way to model the co-occurrence of different objects in the image scene. However, our implementation differs from theirs in the computation of attention scores (we rescale their value to avoid value explosion) and in how the enhanced feature maps are embedded to the original ones (we add them element-wise instead of concatenating them to avoid increasing parameter overhead).
More importantly, we propose \textbf{placing multiple CAB modules} at different network bottlenecks to construct a hierarchical attention mechanism. Indeed, we use CAB on the output of: (i) the V1 layer, representing a coarse and global context, of (ii) the prefrontal cortex (PFC), representing a semantically rich, goal-oriented, context for top-down information flow, and finally of (iii) the IT/LOC layer, which is where the final representation for the object is constructed for recognition.
Lastly, we propose a novel loss term, the \textbf{mini-batch loss}, to push the causality map of samples belonging to the same category closer, so that we foster class-based map alignment. We implement the mini-batch loss as an MSE loss between the causality map of each sample and the average causality map of samples of the same class found in the minibatch during training.

\begin{figure}[tb]
  \centering
  \includegraphics[width=\textwidth]{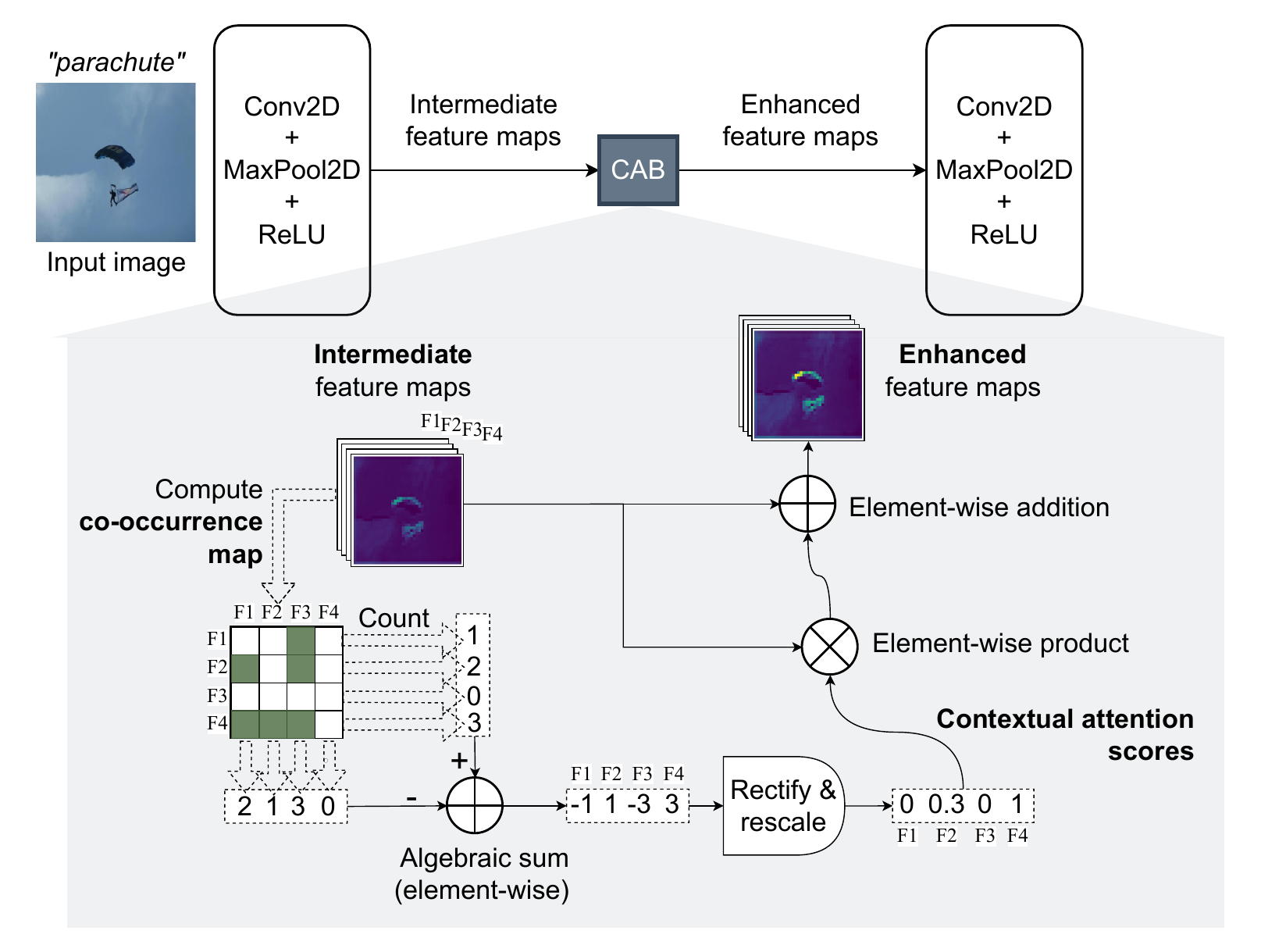}
  \caption{Our Contextual Attention Block (CAB) integrated into a general feed-forward network. As shown, CAB is placed at the convolutional bottleneck of the model. Given intermediate feature maps, the module computes corresponding contextual attention scores through a rectified and rescaled version of the weights obtained from the co-occurrence map.}
  \label{fig:cab_module}
\end{figure}

\subsection{Experimental setup}
We conceive image classification experiments on \textbf{ImagenetteV2 data}, a popular and freely available dataset composed of a subset of 10 easily classified classes from Imagenet (tench, English springer dog, cassette player, chain saw, church, French horn, garbage truck, gas pump, golf ball, and parachute). We utilize the version at 320x320 image resolution. We split the use the official train-val splits and further split the validation set into actual validation and external test set in proportion 60-30. We use the training set for the learning of the models, the validation as internal assessment and hyperparameter tuning, and then test the trained model on the external test data. Evaluation performance metrics were accuracy and F1-score. We repeated the experiments ten times with different random seeds. 
As for the \textbf{total training objective} of our model is a composition of cross-entropy loss, for classification correctness, and MSE loss for class-based causality map alignment.
We compare our CoCoReco model with two ablation versions and one baseline model. To assess the importance of contextual awareness, we remove the CAB module from the network. To assess the importance of bottom-up and top-down modulations (projections) we remove all the skip connections between the different visual areas. Finally, we study the effect of removing the two-branched design, and we train a from-scratch CNN architecture of the same depth as our CoCoReco but with only one branch, representing the traditional bottom-up hierarchy of representations along the ventral stream.


\section{Results and Discussion}
\begin{figure}[tb]
  \centering
  \includegraphics[width=0.75\textwidth]{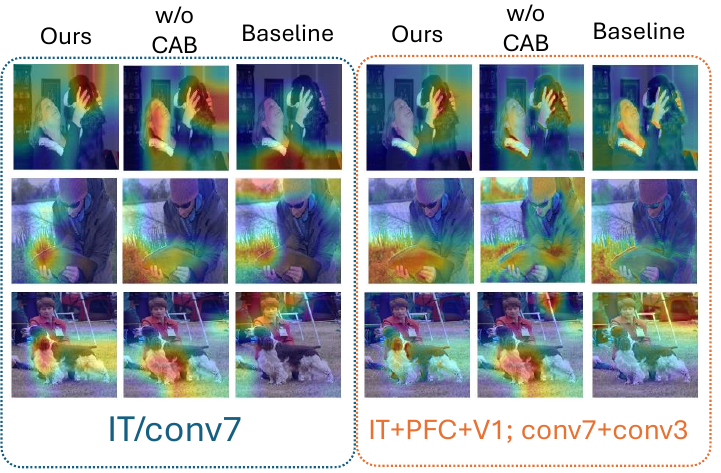}
  \caption{GradCAM activations for some test images. The left panel shows the output when the last convolutional layer before the classifier is chosen as the target layer for the GradCAM computation. The panel on the right shows the output for the same images when a combination of target layers is chosen.}
  \label{fig:gradCAM}
\end{figure}

Table \ref{tab:results} and Fig \ref{fig:gradCAM} summarize our findings. The former shows how our CoCoReco architecture consistently achieves the highest accuracy and F1-score among the investigated models for ten different random seeds. Indeed, we took the mean and standard deviation values across multiple runs and compared CoCoReco to its ablated versions (i.e., without CAB modules and without projections), as well as to the baseline single-branched network. 
\begin{table}[]
    \centering
    \caption{Accuracy and F1-score obtained by our CoCoReco and ablated/baseline models on the external test set of ImagenetteV2 data at resolution 320x320. From left to right, the columns represent our CoCoReco model, its ablated version with no CAB, its ablated version with no bottom-up or top-down projections, and the baseline single-branched CNN. Values are given by mean and standard deviation by training with ten different random seeds.}
    \begin{tabular}{c|c|c|c|c}
       & CoCoReco (ours) & Without CAB & Without projections & Baseline CNN \\ 
       Accuracy & 74.6 (0.63) & 73.8 (0.89) & 73.2 (0.81) & 71.1 (1.0)\\
       F1-score & 74.4 (0.71) & 73.3 (0.91) & 73.1 (0.87) & 71.2 (1.1)\\
    \end{tabular}
    \label{tab:results}
\end{table}
To further assess the benefits of using our proposal, we conducted post-hoc class activation mapping (CAM). We found the explanations produced by our CoCoReco are better than the competing methods. Generally, they are more robust and focused on the salient object of classification without being confounded by confounder aspects of the image that are spuriously associated with the outcome. For instance, Fig \ref{fig:gradCAM} (left panel) shows CoCoReco can focus on the head of the animal while classifying a Springer dog, disregarding attributes of people that frequently coexist in photos of domestic animals. On the same line, our model focuses on the texture of a tench fish while disregarding the features of grass, lake, and trees to arrive at its conclusion.
This holds true even when not only the last convolutional layer before the classifier (i.e., \textit{inferotemporal (IT)} for CoCoReco; \textit{conv7} for the baseline), but also other inner layers are used as targets to compute the GradCAM. The right panel of Fig \ref{fig:gradCAM} shows how CoCoReco has learned important representations in the earlier layers corresponding to V1 areas and semantic-rich PFC. Conversely, the poor quality of explanations produced from \textit{conv7} of the baseline model is confirmed when they are produced jointly from \textit{conv7} and \textit{conv3}; the salient features become the hands and face of the woman holding the dog (first row, last column) or the background grass of the tench fish (second row, last column.)


\section*{Acknowledgements}
This study has received funding from the EU Horizon 2020 program (No 952159, ProCAncer-I) and the Tuscany PAR-FAS PRAMA. The funders had no role in the study's design, data collection, analysis, or manuscript writing.

\bibliographystyle{splncs04}
\bibliography{main}

\begin{thebibliography}{10}
\providecommand{\url}[1]{\texttt{#1}}
\providecommand{\urlprefix}{URL }
\providecommand{\doi}[1]{https://doi.org/#1}

\bibitem{bar2003cortical}
Bar, M.: A cortical mechanism for triggering top-down facilitation in visual object recognition. Journal of cognitive neuroscience  \textbf{15}(4),  600--609 (2003). \doi{https://doi.org/10.1162/089892903321662976}

\bibitem{bar2007proactive}
Bar, M.: The proactive brain: using analogies and associations to generate predictions. Trends in cognitive sciences  \textbf{11}(7),  280--289 (2007). \doi{https://doi.org/10.1016/j.tics.2007.05.005}

\bibitem{bar2006top}
Bar, M., Kassam, K.S., Ghuman, A.S., Boshyan, J., Schmid, A.M., Dale, A.M., H{\"a}m{\"a}l{\"a}inen, M.S., Marinkovic, K., Schacter, D.L., Rosen, B.R., et~al.: Top-down facilitation of visual recognition. Proceedings of the national academy of sciences  \textbf{103}(2),  449--454 (2006). \doi{https://doi.org/10.1073/pnas.0507062103}

\bibitem{bar2006very}
Bar, M., Neta, M., Linz, H.: Very first impressions. Emotion  \textbf{6}(2), ~269 (2006). \doi{https://psycnet.apa.org/doi/10.1037/1528-3542.6.2.269}

\bibitem{bar2001cortical}
Bar, M., Tootell, R.B., Schacter, D.L., Greve, D.N., Fischl, B., Mendola, J.D., Rosen, B.R., Dale, A.M.: Cortical mechanisms specific to explicit visual object recognition. Neuron  \textbf{29}(2),  529--535 (2001). \doi{https://doi.org/10.1016/S0896-6273(01)00224-0}

\bibitem{bomatter2021pigs}
Bomatter, P., Zhang, M., Karev, D., Madan, S., Tseng, C., Kreiman, G.: When pigs fly: Contextual reasoning in synthetic and natural scenes. In: Proceedings of the IEEE/CVF International Conference on Computer Vision. pp. 255--264 (2021). \doi{https://doi.org/10.1109/ICCV48922.2021.00032}

\bibitem{carloni2024exploiting}
Carloni, G., Colantonio, S.: Exploiting causality signals in medical images: A pilot study with empirical results. Expert Systems with Applications  \textbf{249},  123433 (2024). \doi{https://doi.org/10.1016/j.eswa.2024.123433}

\bibitem{carloni2023causality}
Carloni, G., Pachetti, E., Colantonio, S.: Causality-driven one-shot learning for prostate cancer grading from mri. In: Proceedings of the IEEE/CVF international conference on computer vision. pp. 2616--2624 (2023). \doi{https://doi.org/10.1109/ICCVW60793.2023.00276}

\bibitem{choi2012context}
Choi, M.J., Torralba, A., Willsky, A.S.: Context models and out-of-context objects. Pattern Recognition Letters  \textbf{33}(7),  853--862 (2012). \doi{https://doi.org/10.1016/j.patrec.2011.12.004}

\bibitem{choi2020proposal}
Choi, S.H., Jeong, G., Kim, Y.B., Cho, Z.H.: Proposal for human visual pathway in the extrastriate cortex by fiber tracking method using diffusion-weighted mri. Neuroimage  \textbf{220},  117145 (2020). \doi{https://doi.org/10.1016/j.neuroimage.2020.117145}

\bibitem{ding2022efficient}
Ding, Z., Ren, X., David, E., Vo, M., Kreiman, G., Zhang, M.: Efficient zero-shot visual search via target and context-aware transformer. arXiv preprint arXiv:2211.13470  (2022). \doi{https://doi.org/10.48550/arXiv.2211.13470}

\bibitem{gattass1981visual}
Gattass, R., Gross, C.G., Sandell, J.H.: Visual topography of v2 in the macaque. Journal of Comparative Neurology  \textbf{201}(4),  519--539 (1981). \doi{https://doi.org/10.1002/cne.902010405}

\bibitem{gattass1988visuotopic}
Gattass, R., Sousa, A., Gross, C.: Visuotopic organization and extent of v3 and v4 of the macaque. Journal of neuroscience  \textbf{8}(6),  1831--1845 (1988). \doi{https://doi.org/10.1523/JNEUROSCI.08-06-01831.1988}

\bibitem{goodale2013sight}
Goodale, M., Milner, D.: Sight unseen: An exploration of conscious and unconscious vision. OUP Oxford (2013). \doi{https://psycnet.apa.org/doi/10.1093/acprof:oso/9780199596966.001.0001}

\bibitem{goodale1992separate}
Goodale, M.A., Milner, A.D.: Separate visual pathways for perception and action. Trends in neurosciences  \textbf{15}(1),  20--25 (1992). \doi{https://doi.org/10.1016/0166-2236(92)90344-8}

\bibitem{konen2008two}
Konen, C.S., Kastner, S.: Two hierarchically organized neural systems for object information in human visual cortex. Nature neuroscience  \textbf{11}(2),  224--231 (2008). \doi{https://doi.org/10.1038/nn2036}

\bibitem{lehky2007comparison}
Lehky, S.R., Sereno, A.B.: Comparison of shape encoding in primate dorsal and ventral visual pathways. Journal of neurophysiology  \textbf{97}(1),  307--319 (2007). \doi{https://doi.org/10.1152/jn.00168.2006}

\bibitem{mack2011object}
Mack, S.C., Eckstein, M.P.: Object co-occurrence serves as a contextual cue to guide and facilitate visual search in a natural viewing environment. Journal of vision  \textbf{11}(9), ~9--9 (2011). \doi{https://doi.org/10.1167/11.9.9}

\bibitem{milleret2018beyond}
Milleret, C., Bui~Quoc, E.: Beyond rehabilitation of acuity, ocular alignment, and binocularity in infantile strabismus. Frontiers in systems neuroscience  \textbf{12}, ~29 (2018). \doi{https://doi.org/10.3389/fnsys.2018.00029}

\bibitem{oliva2007role}
Oliva, A., Torralba, A.: The role of context in object recognition. Trends in cognitive sciences  \textbf{11}(12),  520--527 (2007). \doi{https://doi.org/10.1016/j.tics.2007.09.009}

\bibitem{paneri2017top}
Paneri, S., Gregoriou, G.G.: Top-down control of visual attention by the prefrontal cortex. functional specialization and long-range interactions. Frontiers in neuroscience  \textbf{11}, ~545 (2017). \doi{https://doi.org/10.3389/fnins.2017.00545}

\bibitem{perry2014feature}
Perry, C.J., Fallah, M.: Feature integration and object representations along the dorsal stream visual hierarchy. Frontiers in computational neuroscience  \textbf{8}, ~84 (2014). \doi{https://doi.org/10.3389/fncom.2014.00084}

\bibitem{rolls2023multiple}
Rolls, E.T., Deco, G., Huang, C.C., Feng, J.: Multiple cortical visual streams in humans. Cerebral Cortex  \textbf{33}(7),  3319--3349 (2023). \doi{https://doi.org/10.1093/cercor/bhac276}

\bibitem{rousselet2004parallel}
Rousselet, G.A., Thorpe, S.J., Fabre-Thorpe, M.: How parallel is visual processing in the ventral pathway? Trends in cognitive sciences  \textbf{8}(8),  363--370 (2004). \doi{https://doi.org/10.1016/j.tics.2004.06.003}

\bibitem{schrimpf2018brain}
Schrimpf, M., Kubilius, J., Hong, H., Majaj, N.J., Rajalingham, R., Issa, E.B., Kar, K., Bashivan, P., Prescott-Roy, J., Geiger, F., et~al.: Brain-score: Which artificial neural network for object recognition is most brain-like? BioRxiv p. 407007 (2018). \doi{https://doi.org/10.1101/407007}

\bibitem{takemura2016major}
Takemura, H., Rokem, A., Winawer, J., Yeatman, J.D., Wandell, B.A., Pestilli, F.: A major human white matter pathway between dorsal and ventral visual cortex. Cerebral cortex  \textbf{26}(5),  2205--2214 (2016). \doi{https://doi.org/10.1093/cercor/bhv064}

\bibitem{torralba2003contextbased}
Torralba, Murphy, Freeman, Rubin: Context-based vision system for place and object recognition. In: Proceedings Ninth IEEE International Conference on Computer Vision. pp. 273--280 vol.1 (2003). \doi{10.1109/ICCV.2003.1238354}

\bibitem{whitwell2014two}
Whitwell, R.L., Milner, A.D., Goodale, M.A.: The two visual systems hypothesis: new challenges and insights from visual form agnosic patient df. Frontiers in neurology  \textbf{5}, ~255 (2014). \doi{https://doi.org/10.3389/fneur.2014.00255}

\bibitem{zhang2020putting}
Zhang, M., Tseng, C., Kreiman, G.: Putting visual object recognition in context. In: Proceedings of the IEEE/CVF conference on computer vision and pattern recognition. pp. 12985--12994 (2020). \doi{https://doi.org/10.1109/CVPR42600.2020.01300}

\end{thebibliography}
\end{document}